%% file: main.tex
\documentclass[10pt,twocolumn,letterpaper]{article}

\usepackage{cvpr}              
\usepackage{graphicx}
\usepackage{amsmath}
\usepackage{amssymb}
\usepackage{booktabs}
\usepackage{multirow} 
\usepackage{makecell}
\usepackage{xcolor}
\usepackage{tabularx}
\usepackage{placeins}
\input{preamble}
\definecolor{cvprblue}{rgb}{0.21,0.49,0.74}
\usepackage[pagebackref,breaklinks,colorlinks,allcolors=cvprblue]{hyperref}

\def\paperID{18634} 
\def\confName{CVPR}
\def\confYear{2026}

\title{Turning Generators into Retrievers: Unlocking MLLMs for Natural Language-Guided Geo-Localization}
\author{
     Yuqi Chen$^{1}$\quad Xiaohan Zhang$^{1}$\quad Ahmad Arrabi$^{1}$\quad Waqas Sultani$^{2}$\quad Chen Chen$^{3}$ \quad Safwan Wshah$^{1*}$ \\[0.5em]
    $^{1}$Vermont Artificial Intelligence Lab, Department of Computer Science, University of Vermont \\
    $^{2}$Intelligent Machines Lab, Department of Artificial Intelligence, Information Technology University \\
    $^{3}$Institute of Artificial Intelligence, University of Central Florida \\[.5em]
    \small \quad $^{*}$Corresponding and senior author.
}
\begin{document}
\maketitle
\input{sec/0_abstract}    
\vspace{-1.5em}
\input{sec/1_intro}

\input{sec/2_related}
\input{sec/3_method}
\input{sec/4_experiment}
\input{sec/5_conclusion}

\section*{Acknowledgments}
This work was partially supported by the National Science Foundation under Grant No. 2218063.


{
    \small
    \bibliographystyle{ieeenat_fullname}
    \bibliography{main}
}
\end{document}

%% file: sec/0_abstract.tex
\begin{abstract}
Natural-language Guided Cross-view Geo-localization (NGCG) aims to retrieve geo-tagged satellite imagery using textual descriptions of ground scenes. While recent NGCG methods commonly rely on CLIP-style dual-encoder architectures, they often suffer from weak cross-modal generalization and require complex architectural designs. In contrast, Multimodal Large Language Models (MLLMs) offer powerful semantic reasoning capabilities but are not directly optimized for retrieval tasks. In this work, we present a simple yet effective framework to adapt MLLMs for NGCG via parameter-efficient finetuning. Our approach optimizes latent representations within the MLLM while preserving its pretrained multimodal knowledge, enabling strong cross-modal alignment without redesigning model architectures. Through systematic analysis of diverse variables, from model backbone to feature aggregation, we provide practical and generalizable insights for leveraging MLLMs in NGCG. Our method achieves SOTA on GeoText-1652 with a 12.2\% improvement in Text-to-Image Recall@1 and secures top performance in 5 out of 12 subtasks on CVG-Text, all while surpassing baselines with far fewer trainable parameters. These results position MLLMs as a robust foundation for semantic cross-view retrieval and pave the way for MLLM-based NGCG to be adopted as a scalable, powerful alternative to traditional dual-encoder designs. Project page and code are available at \url{https://yuqichen888.github.io/NGCG-MLLMs-web/}.
\end{abstract}

%% file: sec/1_intro.tex
\section{Introduction}
\label{sec:intro}


Cross-View Geo-Localization (CVGL)~\cite{siamese_cnn_cvgl,cvusa,vigor,cvgl,university1652} has become a foundational task in geo-spatial learning, as it learns to predict the location of a query ground or drone view image by matching against a database of geo-tagged satellite/aerial observations, enabling many downstream tasks such as drone navigation, autonomous driving, and augmented reality~\cite{survey}. However, existing CVGL methods typically assume that visual ground input is available, overlooking scenarios where only textual descriptions of a location can be provided. For example, emergency calls often rely solely on verbal descriptions rather than photographs or videos. To address this limitation, recent studies \cite{geotext, cvgtext} introduce \textbf{N}atural-language \textbf{G}uided \textbf{C}ross-view \textbf{G}eo-localization (\textbf{NGCG}), which formulates the task as a cross-modal retrieval problem: given a textual description of a ground- or drone-view scene, the goal is to retrieve the corresponding location from a geo-referenced satellite imagery. This paradigm enables CVGL in situations where visual data is unavailable, incomplete, or costly to obtain. Most importantly, NGCG supports practical applications in large-scale mapping platforms (e.g., Google Maps and Bing Maps), where robust alignment between language and geographic imagery is crucial for natural language-guided search, navigation, and scene understanding \cite{robot_nav, gemini, toolformer}.

\begin{figure}[!t] 
    \centering
    \includegraphics[width=.85\linewidth]{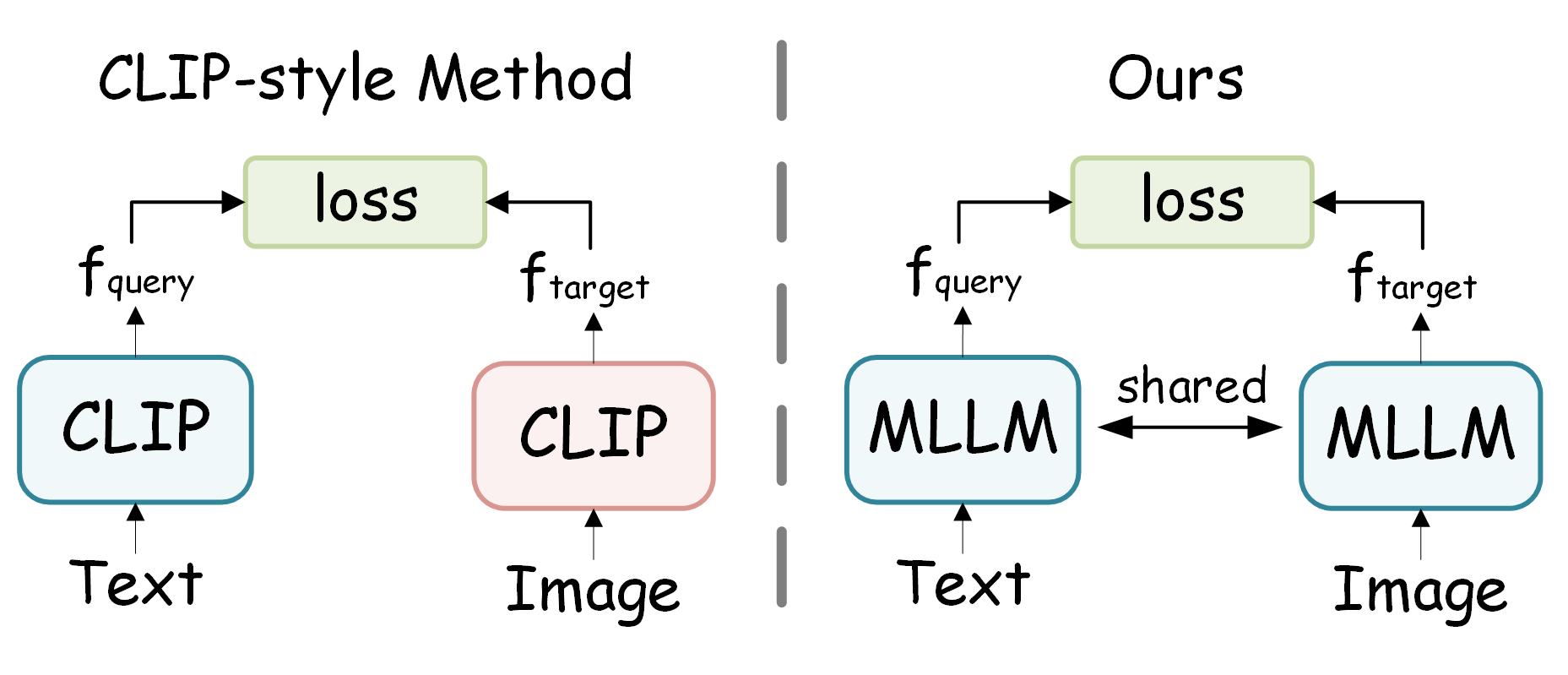} 
    \caption{Comparison between the CLIP-style Dual-Encoder Architecture and Our Unified MLLM Framework.}
    \label{fig:clip_vs_mllm}
    \vspace{-28pt}
\end{figure}

Although NGCG is a highly practical and impactful direction, only a limited number of studies have explored this problem to date~\cite{geotext, cvgtext}. Both GeoText~\cite{geotext} and CrossText2Loc~\cite{cvgtext} adopt CLIP-style Vision-Language Models (VLMs)~\cite{clip}, which use two separate encoders for images and text to learn alignments through contrastive objectives~\cite{simclr, infonce}. This design is a natural fit for cross-modal retrieval, since it allows the two modalities to be projected into a shared embedding space. However, dual-encoder architectures fundamentally limit cross-modal generalization~\cite{defense_dual}, since they defer modality interaction to only the final embedding stage. Similar findings have been reported in CVGL~\cite{sample4geo}, where models that share representations across viewpoints consistently outperform dual-encoder baselines. Furthermore, existing NGCG methods must introduce carefully engineered alignment modules and heavily tuned loss terms~\cite{geotext} to partially compensate for their architectural bottleneck, indicating that dual-encoder VLMs are inherently inefficient at modeling the fine-grained spatial correspondence required for NGCG.

\begin{table}[t]
\centering
\caption{Classification of Research Questions (RQs) for adapting MLLMs to NGCG, categorized by key aspects: Architecture (Arch.), Adaptation (Adap.), and Optimization (Opt.).}
\label{tab:RQs}
\begin{tabular}{p{.8cm} p{7cm}}
\toprule
\textbf{Aspect} & \textbf{Research Question} \\
\midrule

Arch. &(RQ1)\;How does the parameter scale of the MLLM backbones affect NGCG performance? \\
      &(RQ2)\;What is the best way to aggregate sequence embeddings to capture cross-view scene semantics? \\
\midrule

Adap. &(RQ3)\;Which tuning strategy is optimal? Specifically, Full Fine-Tuning (FFT) versus LoRA-based adaptation? \\
      &(RQ4)\;How do the hyperparameter configurations of the parameter-efficient module (e.g., LoRA) influence adaptation performance? \\
\midrule

Opt. &(RQ5)\;How do design choices within the contrastive learning objective influence the discriminative capability and stability of the learned embeddings? \\
\bottomrule
\end{tabular}
\vspace{-17pt}
\end{table}


In contrast to CLIP-style VLMs, recent advances in Multimodal Large Language Models (MLLMs)~\cite{gemini, gpt4, blip2, llava} have demonstrated remarkable capabilities in cross-modal reasoning and instruction following. By \textit{jointly} processing visual and textual inputs within a unified architecture, MLLMs can capture fine-grained semantic correspondence and contextual relationships that are difficult for dual-encoder models to express. This indicates that MLLMs can better support NGCG, which relies on connecting descriptive scene language to visual observations in a coherent manner. The key difference between this approach and existing CLIP-style methods is illustrated in \Cref{fig:clip_vs_mllm}. However, directly applying MLLMs to NGCG is not straightforward. First, most MLLMs are optimized for generation tasks rather than retrieval, producing textual outputs rather than discriminative embeddings suitable for retrieval tasks. Second, the training objectives used in MLLMs, such as next-token prediction~\cite{att_need} and RLHF optimization~\cite{instructgpt}, are not aligned with the retrieval objective in NGCG, which typically requires contrastive learning. As a result, the learned representations are not naturally structured for cross-modal matching. To validate this, we performed a zero-shot evaluation of the pretrained InternVL3.5-1B~\cite{internvl3.5} on the CVG-Text~\cite{cvgtext} New York subset. The model's R@1 score is only 0.92\%, a result negligibly better than random guessing, which confirms the necessity of task-specific adaptation. The core challenge, therefore, is not simply using MLLMs, but understanding how to reshape their latent space for geographically grounded cross-modal matching while preserving their reasoning capabilities.

Motivated by this finding and the recent efforts to adapt MLLMs for representation learning~\cite{vlm2vec,exploring_mllms,llm2clip}, we take a step toward bringing MLLMs to the challenging NGCG task. Note that CrossText2Loc~\cite{cvgtext} leveraged MLLMs such as GPT-4o for confidence re-ranking. Since relying on closed-source APIs, this design is costly, lacks flexibility, and is difficult to fine-tune. In contrast, we are the first to explore MLLMs as an end-to-end trainable feature extractor for directly tackling the NGCG task, thereby eliminating the need for external API calls. To uncover how MLLMs can be most effectively adapted for NGCG, we conduct a comprehensive study guided by five central research questions (\textbf{RQs}) in \Cref{tab:RQs}. Through systematic experiments addressing these questions, we provide actionable insights into which adaptation factors matter most for NGCG. Our approach achieves state-of-the-art performance on GeoText-1652~\cite{geotext}, improving Text-to-Image retrieval accuracy by $+12.2\%$ at R@1, and delivers strong results on CVG-Text~\cite{cvgtext}. Overall, our findings highlight that parameter-efficient representation adaptation offers a scalable and effective direction for NGCG, simplifying model design while advancing performance. The main contributions of this paper are as follows.
\begin{enumerate}
    \item To the best of our knowledge, we are the first to explore MLLMs for the Natural-language Guided Cross-view Geo-localization (NGCG). We study an end-to-end parameter-efficient adaptation approach that preserves the conversation and reasoning ability of MLLMs, while enabling discriminative cross-modal retrieval.

    \item We systematically explore key factors, including backbone scale, feature aggregation, and fine-tuning configurations. And we reveal practical insights and best practices for effectively adapting MLLMs to the NGCG task.

    \item Our approach achieves a Recall@1 of 25.8\% on GeoText-1652 and 5 best out of 12 subtasks on CVG-Text, outperforming prior NGCG methods and demonstrating the effectiveness of parameter-efficient MLLM adaptation. 
\end{enumerate}

%% file: sec/2_related.tex
\section{Related Work}
\label{sec:related_work}

\textbf{Cross-View Geo-localization} aims to match ground or drone-view images with satellite imagery, often formulated as a retrieval problem~\cite{survey}. Early methods primarily relied on CNN-based architectures~\cite{hu2018cvm, shi2020optimal, siamese_cnn_cvgl}, which struggle with large viewpoint variations. Transformer-based approaches~\cite{SAIG, geodtr, zhu2022transgeo,geodtr+,zhang2023cross,cvgl_cross_layer:yang} have since leveraged self-attention~\cite{att_need} to enhance viewpoint robustness. Representation quality has been further improved using geometric priors such as polar coordinate transformations~\cite{spatial_aggregation:shi, ye2024cross} or refined with hard negative mining~\cite{sample4geo, cai2019ground} to increase discriminability in the embedding space. Furthermore, generative models have been employed for data augmentation and domain adaptation across viewpoints~\cite{cvgl_condition:regmi, cvgl_panorma_synthesis:wu, cvgl_diffusion:arrabi,zhang2025vici}. 

Recently, Natural-language Guided Cross-view Geolocalization (NGCG) has emerged as a new research direction. GeoText-1652~\cite{geotext} and CVG-Text~\cite{cvgtext} demonstrated the feasibility of geo-localization from textual scene descriptions. However, existing NGCG approaches~\cite{geotext, cvgtext} rely on CLIP-style models, which encode images and texts using separate encoders and perform matching in a shared embedding space. While effective for general cross-modal retrieval, such dual-encoder architectures offer weak cross-modal interaction and result in task-specific models. As a result, these methods often struggle to align textual descriptions with fine-grained geographic visual cues.



In this paper, we investigate how Multimodal Large Language Models (MLLMs) can effectively provide richer cross-modal cues through their inherent multimodal design and large-scale pretraining data. Our goal is to design a parameter-efficient framework that leverages the multimodal understanding embedded in MLLMs to address the challenges of NGCG, moving beyond existing CLIP-based methods~\cite{cvgtext,geotext} and intricate loss design~\cite{geotext}.\\

 \vspace{-8pt}
\noindent\textbf{Multimodal Large Language Models} (MLLMs)~\cite{smolvlm,internvl3.5,qwen2,phi} have recently demonstrated strong visual–language understanding across domains such as remote sensing~\cite{mm_rs1,mm_rs2}, autonomous driving~\cite{mm_driving1,mm_driving2}, and medical imaging~\cite{mm_medical1,mm_medical2}. Their success stems from large-scale multimodal pretraining and understanding. However, directly applying MLLMs to NGCG is far from straightforward. NGCG requires fine-grained, discriminative alignment between textual descriptions and geographically grounded visual cues, whereas MLLMs are optimized for next-token prediction and RLHF-guided instruction following. These objectives produce embeddings that are semantically expressive but not geometrically structured nor metrically aligned for contrastive retrieval. Consequently, off-the-shelf MLLM representations are often unsuitable for NGCG despite their high-level multimodal understanding.

To address this gap, we investigate how to purposefully adapt MLLMs so that their rich multimodal priors become useful for NGCG. We conduct a comprehensive empirical study on parameter-efficient fine-tuning strategies that preserve pretrained knowledge while reshaping the embedding space for discriminative cross-modal alignment. Our study systematically explores key design elements, including backbone choice, feature aggregation mechanisms and LoRA scaling, and provides concrete insights into how MLLMs can be effectively transformed into strong NGCG models. This work represents an initial step toward leveraging MLLMs’ multimodal reasoning for the NGCG task.



%% file: sec/3_method.tex
\vspace{-2pt}
\section{Methodology}\label{sec:method}

\begin{figure}[!t] 
\vspace{-10pt}
    \centering
    \includegraphics[width=.99\linewidth]{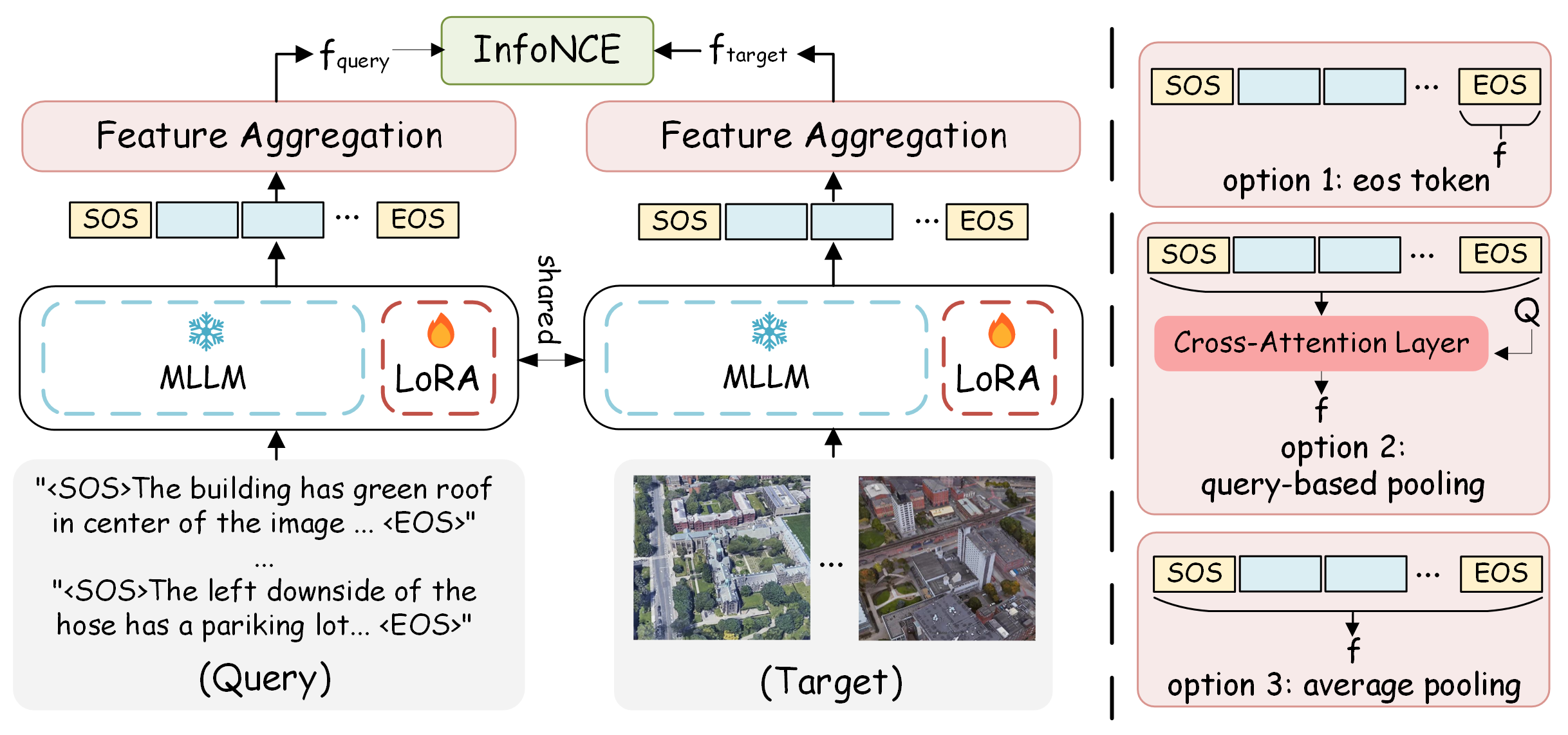} 
    \caption{The framework leverages a pre-trained MLLM for feature encoding, processing both text and visual data via its shared weight design. The adaptation framework relies on LoRA, combined with an InfoNCE loss function, to align the MLLM's internal representation space for the NGCG retrieval task.}
    \label{fig:framework}
    \vspace{-15pt}
\end{figure}

\subsection{Problem Formulation}
The task of Natural-language Guided Cross-view Geo-localization (NGCG) can be formally defined as a retrieval task. Specifically, given a query text $q$ describing a ground-level scene (e.g., a street-view image), the objective is to retrieve its geographically corresponding satellite image $s^m$, where $m$ denote index of the matching satellite image from a database $\mathcal{S} = \{ s^1, s^2, \dots, s^N \}$ which contains $N$ geo-tagged satellite image candidates.

This process can be formulated as learning a feature encoder $G_\theta$, where $\theta$ represents the trainable parameters, that encodes $q$ and $s^{i \in \{1,\dots, N\}}$ into latent features such that
\begin{equation}
    m = \underset{i\in\{1, \dots, N\}}{\arg\min}\, d(\mathbf{f}_{q}, \mathbf{f}^i_s),
\end{equation}
where $d(\cdot, \cdot)$ denotes a latent space distance (i.e., $L_2$ distance). $\mathbf{f}_{q}$ and $\mathbf{f}_s^i$ represent the feature representation of the query and satellite images, respectively.\\

\vspace{-1.em}
\subsection{Adapting MLLMs to NGCG}
\subsubsection{MLLM Backbone and Feature Aggregation}
\label{subsubsec:backbone}
Accurate NGCG requires embeddings that capture both semantic content and spatial relationships across two views. To achieve this, we utilize a Multimodal Large Language Model (MLLM) as our feature encoder, $G_\theta$, which extracts the last layer's hidden states that can be aggregated into discriminative representations. An overview of our framework is shown  in~\Cref{fig:framework}. We consider MLLM backbones of varying scale, from SmolVLM-256M~\cite{smolvlm} to InternVL3.5
-1B~\cite{internvl3.5}, to understand how model scale influences the quality of cross-view embeddings, which is the core of \textbf{RQ1}. Besides backbones, extracting feature representation is also necessary to drive NGCG. Thereby, we explore 3 different feature aggregation strategies that aggregate the last layer's hidden states into embedding features to answer \textbf{RQ2}. A comparison between the above-mentioned three different feature aggregation strategies is illustrated on ~\Cref{fig:framework}.

\noindent1) \textit{[EOS] Token Pooling:} Following common practice in representation extraction for autoregressive models \cite{simcse, llm2vec, vlm2vec}, we take the last token (i.e., the [EOS] token) from the final hidden state as the feature representation for both reference satellite image and query text inputs.

\noindent2) \textit{Query-based Pooling}: Leveraged by prominent models, such as BLIP~\cite{blip2}, DETR~\cite{query:detr}, and Perciever families~\cite{query:perceiver}, which employs one or more learned query vectors to compute attention over input features, thereby distilling the features into a compact representation by emphasizing elements most relevant to the task. We use the single learnable query $Q \in \mathbb{R}^{1 \times d}$, where $d$ is the MLLM's latent dimension.

\noindent3) \textit{Average Pooling:} We also employ average pooling that aggregates all valid tokens in the last hidden states as a single feature representation. 


\vspace{-2.5pt}
\subsubsection{Model Fine-tuning}
\label{subsubsec:lora}
While large-scale MLLMs encode rich multimodal knowledge, task-specific adaptation is necessary to optimize embeddings for NGCG. Full fine-tuning updates all model parameters and can potentially achieve higher retrieval accuracy due to the large number of trainable weights. However, it comes with two main drawbacks: it is computationally expensive, and it risks overwriting pre-trained knowledge, including the model’s conversational and reasoning capabilities that we aim to preserve. Therefore, we employ the widely adopted Low-Rank Adaptation (LoRA)\cite{lora} which freezes the original pretrained weight $W_{0} \in \mathbb{R}^{d \times k}$, where $d$ is the input dimension and $k$ is the output dimension, and injects two trainable rank decomposition matrices $B \in \mathbb{R}^{d \times r}$ and $A \in \mathbb{R}^{r \times k}$, where the low rank $r \ll \min(d, k)$. Specifically, for a given layer with input vector $x$, the original weight matrix $W_0$ is updated by $\Delta W$. This update is approximated by the scaled low-rank product
\begin{equation}
    \Delta W = \frac{\alpha}{r} (B A),
\end{equation}
where factor $\frac{\alpha}{r}$ controls the magnitude of the update. The resulting modified hidden state $h'$ is then calculated as the sum of the original output and this low-rank update
\begin{equation}
h' = W_{0}x + \frac{\alpha}{r} \left( B A\right) x.
\end{equation}
We used a full-coverage strategy, applying LoRA modules to both the multi-head attention matrices and the intermediary Multi-Layer Perceptron (MLP) within MLLMs. 

Although LoRA-based adaptation significantly reduces the number of trainable parameters while maintaining pre-trained abilities, the limited capacity can make optimization difficult, particularly for tasks like NGCG. To systematically explore these trade-offs, we first compare full fine-tuning vs. LoRA adaptation (\Cref{subsec:fine-tuning}) to assess how the adaptation method impacts both performance and efficiency (\textbf{RQ3}). We then vary LoRA scaling parameters (\Cref{subsec:lora}) to understand how low-rank update strength affects convergence and retrieval quality (\textbf{RQ4}). This study provides guidance for balancing efficiency, adaptation strength, and preservation of pre-trained knowledge in NGCG. 


\subsubsection{Training objective}\label{subsec:objective}
To train our framework on the NGCG tasks, we follow prior arts~\cite{geotext,sample4geo,ye2024cross,cvgtext} and adopt the well-known InfoNCE~\cite{infonce} loss to optimize the distance between positive and negative pairs. Specifically, assuming a mini-batch of satellite-text pairs, each text query $\mathbf{f}_{q}$ in this mini-batch should have one positive corresponding satellite image $\mathbf{f}_s^{+}$. The goal is to encourage alignment with its corresponding satellite embedding $\mathbf{f}_s^+$ while being contrasted against negatives that are all the other satellite images in this mini-batch. Mathematically, the loss is defined as
\begin{equation}
\mathcal{L}_{\text{InfoNCE}} = - \log \frac{\exp \left( \mathbf{f}_{q} \cdot \mathbf{f}_{s}^{+} / \tau \right) }{\sum_{i=0}^{B} \exp \left( \mathbf{f}_{q} \cdot \mathbf{f}_{s}^{i} / \tau \right) },
\label{eq:info_nce_loss}
\end{equation}
where $\tau$ refers to the temperature parameter, and ($\cdot$) denotes dot product between L2-normalized embeddings. In our \textbf{RQ5}, we investigate the effect of $\tau$ during the training phase. Specifically, we compare different values of $\tau$ or make it learnable. Interestingly, contrary to the common practice of learnable $\tau$~\cite{sample4geo,ye2024cross}, we find that a fixed temperature not only stabilizes training and produces more discriminative embeddings for NGCG retrieval, highlighting an important design consideration for contrastive learning in this domain.

%% file: sec/4_experiment.tex
 \vspace{-5pt}
\section{Experiment}\label{sec:experiment}

\textbf{Experiment Setup.} We conducted our experiment on two recently proposed datasets, CVG-Text~\cite{cvgtext} and Geo-Text~\cite{geotext}. Following the prior works \cite{cvusa,cvact,vigor,cvgtext,geotext}, we use the Top-K Recall accuracy (R@K) to assess retrieval localization performance. Specifically, given a query text, if its ground-truth satellite image is within the Top-K retrieved results, it counts as successful localization. Additionally, following the benchmark \cite{cvgtext}, we also use the localization recall rate (L@D), which refers to the proportion of retrieved results where the distance between the ground truth satellite location and Top-1 predicted satellite location is below a certain threshold D. \textit{For more implementation details, please refer to our supplementary materials.}


\begin{table}[]
\vspace{-5pt}
\footnotesize
\centering
\setlength{\tabcolsep}{4.1pt} 
\caption{\textbf{Comparison of natural-language guided cross-view geo-localization performance on the GeoText-1652 dataset.} We report recall rates (\%) at different top-K retrieval thresholds. R@K represents the Top-K recall rate. \#TPs: Number of Trainable Parameters. [Key: \textbf{Best}, \underline{Second Best}]}
\vspace{-8pt}
\label{tab:main_geotext}
\begin{tabular}{c|c|ccc|ccc} 
\toprule
\multirow{2}{*}{\textbf{Method}} & \multirow{2}{*}{\textbf{\makecell{\#TPs}}} & \multicolumn{3}{c|}{\textbf{Text $\to$ Image }} & \multicolumn{3}{c}{\textbf{Image $\to$ Text }} \\
\cmidrule(lr){3-5} \cmidrule(lr){6-8} 
& & \textbf{R@1} & \textbf{R@5} & \textbf{R@10} & \textbf{R@1} & \textbf{R@5} & \textbf{R@10} \\
\midrule
UNITER\cite{chen2020uniter} & 300M & 10.6 & 20.4 & 26.1 & 21.4 & 43.4 & 59.5\\
METER\cite{meter_swin} & 380M & 11.3 & 21.5 & 27.3 & 22.7 & 46.3 & 60.7\\
ALBEF\cite{albef} & 210M & 12.3 & 22.8 & 28.6 & 22.9 & 49.5 & 62.3\\
XVLM\cite{xvlm} & 216M & 13.1 & 23.5 & 29.2 & 23.6 & 50.0 & 63.2\\
GeoText\cite{geotext} & 216M & \underline{13.6} & \underline{24.6} & \underline{31.2} & \underline{26.3} & \underline{53.7} & \underline{66.9} \\
\midrule 
Ours$_{\textit{InternVL}}$& 9M& \textbf{25.8} & \textbf{41.0} & \textbf{49.4} & \textbf{ 34.4} & \textbf{63.3} & \textbf{75.1}\\
\bottomrule
\end{tabular}
\vspace{-8pt}
\end{table}

\begin{table}[t]
  \centering
  \footnotesize
  \caption{\textbf{Performance comparison on the GeoText-1652 24G test set.} We report Recall@K (R@K, \%) for both Text-to-Image (T$\to$I) and Image-to-Text (I$\to$T) retrieval tasks. Bold values indicate the best performance.}
  \label{tab:main_geotext_24g}
  \setlength{\tabcolsep}{3.5pt} 
  \begin{tabular}{l|ccc|ccc}
    \toprule
    \multirow{2}{*}{\textbf{Method}} & \multicolumn{3}{c|}{\textbf{Text $\to$ Image}} & \multicolumn{3}{c}{\textbf{Image $\to$ Text}} \\
    \cmidrule(lr){2-4} \cmidrule(lr){5-7}
    & \textbf{R@1} & \textbf{R@5} & \textbf{R@10} & \textbf{R@1} & \textbf{R@5} & \textbf{R@10} \\
    \midrule
    CLIP-L/14~\cite{clip} & 37.7 & 52.9 & 60.3 & 49.9& 82.0 &90.9 \\
    GeoText~\cite{geotext} & 29.9 & 46.3 & 54.1 & 50.1 & 81.2 & 90.3 \\
    \midrule
    Ours$_{\textit{InternVL}}$  & \textbf{45.7} & \textbf{62.9} & \textbf{70.3} & \textbf{58.3} & \textbf{85.9} & \textbf{92.9}\\
    \bottomrule
  \end{tabular}
\end{table}

\begin{table*}[]
\vspace{-10pt}
\centering
\footnotesize
\setlength{\tabcolsep}{5.0pt} 
\caption{\textbf{Comparison of natural-language guided cross-view geo-localization performance on the CVG-Text dataset for Text-to-Satellite Image and Text-to-OSM Image Retrieval.} We report recall rates (\%) and localization hit rates (\%) based on thresholded retrieval. R@1 represents Top-1 Recall (the correct candidate is the top result). L@50 represents the localization recall rate, where the correct candidate is retrieved and the predicted location error is less than 50 meters. [Key: \textbf{Best}, \underline{Second Best}]}
\label{tab:main_cvgtext}
\begin{tabular}{c|c|cccccc|cccccc}
\toprule
\multirow{3}{*}{\textbf{Method}} & \multirow{3}{*}{\textbf{\makecell{\# Trainable \\ Params}}} & \multicolumn{6}{c|}{\textbf{Ground Text $\to$ Satellite Image}} & \multicolumn{6}{c}{ \textbf{Ground Text $\to$ OSM Image}}\\
\cmidrule(lr){3-8} \cmidrule(lr){9-14}
& & \multicolumn{2}{c}{\textbf{NewYork}} & \multicolumn{2}{c}{\textbf{Brisbane}} & \multicolumn{2}{c|}{\textbf{Tokyo}} & \multicolumn{2}{c}{\textbf{NewYork}} & \multicolumn{2}{c}{\textbf{Brisbane}} & \multicolumn{2}{c}{\textbf{Tokyo}} \\
\cmidrule(lr){3-4} \cmidrule(lr){5-6} \cmidrule(lr){7-8} \cmidrule(lr){9-10} \cmidrule(lr){11-12} \cmidrule(lr){13-14}
& & \textbf{R@1} & \textbf{L@50} & \textbf{R@1} & \textbf{L@50} & \textbf{R@1} & \textbf{L@50} & \textbf{R@1} & \textbf{L@50} & \textbf{R@1} & \textbf{L@50} & \textbf{R@1} & \textbf{L@50} \\
\midrule
ViLT\cite{kim2021vilt} & 87.4M & 11.58 & 15.58 & 11.00 & 14.50 & 10.83 & 15.50 & 5.83 & 9.92 & 8.67 & 11.75 & 4.67 & 9.17 \\
X-VLM\cite{xvlm} & 216M & 15.74 & 16.86 & 15.67 & 17.60 & 12.46 & 14.34 & 16.14 & 17.26 & 20.46 & 21.94 & 9.53 & 10.94 \\
SigLIP-B/16\cite{zhai2023sigmoid} & 203M & 19.67 & 21.08 & 19.58 & 22.00 & 15.58 & 17.92 & 20.17 & 21.58 & 25.58 & 27.42 & 11.92 & 13.67 \\
CLIP-L/14\cite{clip} & 428M & 35.08 & 37.08 & 34.08 & 37.25 & 28.08 & 30.50 & 31.50 & 33.58 & 32.50 & 34.67 & 21.00 & 23.17 \\
BLIP\cite{li2022blip} & 470M & 34.58 & 37.25 & 34.50 & 38.17 & 29.75 & 33.67 & 52.92 & 55.92 & \underline{43.00} & 46.33 & 30.67 & 34.50 \\
CrossText2Loc\cite{cvgtext} & 428M & {\textbf{46.25}} & {\textbf{48.75}} & {\underline{43.58}} & {\underline{47.42}} & {\textbf{36.83}} & {\textbf{39.58}} & {\textbf{59.08}} & {\underline{62.00}} & {\textbf{46.08}} & {\textbf{48.67}} & {\underline{34.33}} & {\underline{38.33}} \\
\midrule
Ours$_{\textit{InternVL}}$ & 9M  & \underline{44.75} & \underline{47.50} & \textbf{45.25}& \textbf{49.00} & \underline{33.00} & \underline{36.50} & \underline{56.92} &\textbf{63.33} & 37.17 & 41.83 & \textbf{39.17} & \textbf{44.33} \\

\bottomrule
\end{tabular}
\vspace{-15pt}
\end{table*}

\subsection{Main Results}\label{subsec:result}

\textbf{GeoText-1652.} 
\Cref{tab:main_geotext} presents the performance of our models compared to prior state-of-the-art(SOTA) methods on the GeoText-1652~\cite{geotext} benchmark. The results of our approach demonstrate superiority over all existing methods, setting a new SOTA across all evaluation metrics. For Text-to-Image retrieval, our method achieves $25.8\%$ at R@1, marking an absolute improvement of $+12.2\%$ points over the baseline ($13.6\%$). Similarly, for Image-to-Text retrieval, our model sets a new SOTA with $34.4\%$ at R@1, confirming its comprehensive strength in both retrieval directions. Note that our model achieves these results using only 9M trainable parameters for the LoRA module, which is significantly parameter-efficient than prior works.

We also evaluate our method on the GeoText-1652 24G subset provide by the original authors in \Cref{tab:main_geotext_24g}. In this setting, Ours-I significantly outperforms CLIP-L/14 (with LoRA), which is the widely recognized SOTA for cross-modal retrieval. The consistent gains validate that our method effectively captures spatial relationships.


\noindent\textbf{CVG-Text.} We benchmark our method on the CVG-Text dataset, as shown in Table \ref{tab:main_cvgtext}. First, our model demonstrates superiority over general CLIP-based baselines, consistently outperforming all of them across all benchmarks. Compared with the prior SOTA, CrossText2Loc~\cite{cvgtext}, our approach achieves comparable performance while utilizing substantially fewer trainable parameters. Overall, our method obtains 5 top-performing results and 5 second-best results out of the 12 total sub-tasks. Specifically, for Text-to-OSM retrieval, we demonstrate superior capabilities, particularly in the Tokyo subset. Our model achieves $39.17\%$ on R@1 (an improvement of $+4.84\%$) and $44.33\%$ on L@50 (an improvement of $+6.00\%$) over the previous SOTA. Second, for the Text-to-Satellite task, we set a new SOTA on the Brisbane dataset, achieving $45.25$ on R@1 (an improvement of $+1.67\%$) and $49.00\%$ on L@50 (an improvement of $+1.58\%$). We acknowledge that our method's performance on Text-to-OSM in Brisbane is lower than that of CrossText2Loc and BLIP. We hypothesize that the lower performance is attributed to the comparatively low building and landmark density in Brisbane, relative to Tokyo and New York, which results in sparse annotations within the Brisbane OSM data. Despite a few underperformances on certain sub-tasks, our approach achieves an outstanding performance while maintaining parameter efficiency.


\section{Understanding Key Design Choices}\label{sec:analysis}
\subsection{Backbone Selection}

We begin by comparing different MLLM backbones to address our first research question \textbf{RQ1}. Specifically, we evaluate three MLLMs of varying scales: InternVL3.5-1B~\cite{internvl3.5}, SmolVLM-500M~\cite{smolvlm}, and SmolVLM-256M~\cite{smolvlm}. Results on CVG-Text benchmarks are summarized in \Cref{tab:backbone_cvgtext}. Overall, reducing the backbone size leads to a moderate drop in NGCG performance. Interestingly, the performance decay is much slower than the reduction in model parameters, suggesting that even small-scale MLLMs retain substantial pretrained knowledge. For Ground Text-to-OSM retrieval task in New York, our method with SmolVLM-256M achieves an R@1 score of $48.67\%$, outperforming comparable-scale models (X-VLM~\cite{xvlm}, SigLip~\cite{zhai2023sigmoid}) and approaching the performance of larger models (BLIP~\cite{blip2}, CLIP~\cite{clip}) (see \Cref{tab:main_cvgtext}). These results suggest that MLLMs already encode rich multimodal knowledge, which can be effectively leveraged through LoRA fine-tuning even when at small model scale.



\noindent\textbf{Takeaway:}  While smaller MLLM backbones lead to some performance drop, the decline is much slower than the reduction in parameters. Even SmolVLM-256M retains strong multimodal knowledge, achieving competitive NGCG performance comparable to large-scale CLIP-based models when fine-tuned with LoRA.

\begin{table}
\centering
\footnotesize
\setlength{\tabcolsep}{3.0pt} 
\caption{Analysis of different scales MLLM backbone on the CVG-Text Benchmark. Metrics R@1 and L@50 are reported. ``G2S'' is short for ``ground text $\rightarrow$ satellite'' and ``G2O'' is short for ``ground text $\rightarrow$ OSM image''. ``S-500M'' and ``S-256M'' are short for ``SmolVLM-500M'' and ``SmolVLM-256M'', respectively. \#TPs: Number of Trainable Parameters. [Key: \textbf{Best}]}
\label{tab:backbone_cvgtext}

\begin{tabular}{c|c|c|cccccc}
\toprule
\multirow{2}{*}{\textbf{Task}}& \multirow{2}{*}{\textbf{Backbone}} & \multirow{2}{*}{\textbf{\makecell{\#TPs}}} & \multicolumn{2}{c}{\textbf{NewYork}} & \multicolumn{2}{c}{\textbf{Brisbane}} & \multicolumn{2}{c}{\textbf{Tokyo}} \\
\cmidrule(lr){4-5} \cmidrule(lr){6-7} \cmidrule(lr){8-9}
& & & \textbf{R@1} & \textbf{L@50} & \textbf{R@1} & \textbf{L@50} & \textbf{R@1} & \textbf{L@50} \\
\midrule
\multirow{3}{*}{G2S} & InternVL & 9M  & \textbf{44.75} & \textbf{47.50} & \textbf{45.25}& \textbf{49.00} & \textbf{33.00} & \textbf{36.50} \\
& S-500M & 6.4M & 36.50 & 39.17 &38.25 & 42.50 & 26.83 & 30.25 \\
& S-256M & 4.1M  & 33.25 & 35.83 & 34.83 &38.17  & 24.17 & 27.17  \\
\midrule
\multirow{3}{*}{G2O} & InternVL & 9M  & \textbf{56.92} &\textbf{63.33} & 37.17 & 41.83 & \textbf{39.17} & \textbf{44.33} \\
& S-500M & 6.4M &    53.92& 60.00 & 40.33 & 45.25 &29.17  & 36.58 \\
& S-256M & 4.1M  &48.67   & 54.33 & \textbf{41.50} & \textbf{46.75} &26.33  &33.17  \\
\bottomrule
\end{tabular}
\end{table}

\begin{table}[!ht]
\vspace{-5pt}
\centering
\footnotesize
\setlength{\tabcolsep}{5pt}
\caption{Analysis on the feature aggregation strategy on Text-to-Satellite task of CVG-Text New York Subset. ``[EOS] is short for [EOS] token pooling. ``Avg.'' is short for average pooling. ``Query'' is short for query-based pooling. [Key: \textbf{Best}]} 
\label{tab:pooling_ablation}
\begin{tabular}{@{}l|cccccc@{}} 
\toprule
\textbf{Pooling} & \textbf{R@1} & \textbf{R@5} & \textbf{R@10} & \textbf{L@50} & \textbf{L@100} & \textbf{L@150} \\
\midrule
\text{[EOS]} & \textbf{44.75} & \textbf{75.92} &\textbf{86.00} &\textbf{47.50} &\textbf{51.33} &\textbf{54.25} \\
Query & 36.83 & 72.50 & 83.58 & 39.17 & 43.00 & 45.33 \\
Avg. & 36.92 & 72.50 & 82.67 & 39.50 & 43.58 & 46.25 \\

\bottomrule
\end{tabular}
 \vspace{-10pt}
\end{table}

\subsection{Feature Aggregation Strategies}\label{subsec:rep}

To answer our \textbf{RQ2}, we compare the three feature aggregation strategies: 1) [EOS] token pooling, 2) query-based pooling, and 3) average pooling, which are detailed in~\Cref{subsubsec:backbone}. Table~\ref{tab:pooling_ablation} reports the performance of these methods. Contrary to the expectation that aggregating all hidden states (as in average or query-based pooling) might yield richer context, the [EOS] token consistently delivers the best results, achieving $39.25\%$ at R@1. Both average pooling and query-based pooling underperform by a noticeable margin across all metrics, even though the latter includes trainable parameters. We attribute this to the fact that the [EOS] token is explicitly supervised during MLLM pretraining to summarize global context, making it inherently well-suited for aggregating global features. In contrast, average and query-based pooling must learn to distill discriminative features from long sequences without prior guidance, which may hinder convergence.

\noindent\textbf{Takeaway:} The [EOS] token provides the most discriminative and stable representation for NGCG, demonstrating that leveraging pretrained summarization tokens is more effective than learning new pooling mechanisms.

\begin{table*}[!t]
\vspace{-15pt}
\footnotesize
\centering
\setlength{\tabcolsep}{3pt} 
\caption{Comparison of fine-tuning strategies on SmolVLM-500M and InternVL3.5-1B for Text-to-Satellite on the CVG-Text Benchmark.}
\label{tab:fine-tuning_satellite}
\begin{tabular}{@{}l|cccc|cccc|cccc@{}} 
\toprule
\multirow{3}{*}{\textbf{backbone}} & \multicolumn{12}{c}{\textbf{Ground Text $\to$ Satellite Image}} \\
\cmidrule(lr){2-13} 
& \multicolumn{4}{c|}{\textbf{New York}} & \multicolumn{4}{c|}{\textbf{Brisbane}} & \multicolumn{4}{c}{\textbf{Tokyo}} \\
\cmidrule(lr){2-5} \cmidrule(lr){6-9} \cmidrule(lr){10-13}
& \textbf{R@1} & \textbf{R@5} & \textbf{L@50} & \textbf{L@100} & \textbf{R@1} & \textbf{R@5} & \textbf{L@50} & \textbf{L@100} & \textbf{R@1} & \textbf{R@5} & \textbf{L@50} & \textbf{L@100} \\
\midrule
SmolVLM-500M (Full) & 20.67 & 54.33 & 22.50 & 25.58 & 21.08 & 52.33 & 23.33 & 26.00 & 13.17 & 43.92 & 16.67 & 20.83 \\
SmolVLM-500M (LoRA) & 36.50 &70.67 & 39.17 & 43.33&38.25 &70.00 & 42.50 &46.25 & 26.83 & 61.17& 30.25 & 35.17 \\
\textbf{Gain} &\small{$+15.83$} &\small{$+16.34$} &\small{$+16.67$} &\small{$+17.75$} &\small{$+17.17$} & \small{$+17.67$} & \small{$+19.17$} &\small{$+20.25$} & \small{$+13.66$} & \small{$+17.25$} & \small{$+13.58$} &\small{$+14.34$} \\
\midrule
InternVL(Full) & 41.58 & 74.75  & 44.17 & 48.17 & 38.67 & 73.58 & 42.92 & 46.00 & 28.83 & 63.83 &32.25 & 37.75 \\
InternVL(LoRA) & 44.75 & 75.92 & 47.50 & 51.33 & 45.25 & 74.92 & 49.00 & 52.17 &33.00 & 67.08 & 36.50 & 40.92\\
\textbf{Gain} & \small{$+3.17$} & \small{$+1.17$}& \small{$+3.33$} &\small{$+3.16$} &\small{$+6.58$}  &\small{$+1.34$}&\small{$+6.08$}&\small{$+6.17$}&\small{$+4.17$} &\small{$+3.25$}&\small{$+4.25$}&\small{$+3.17$} \\
\bottomrule
\end{tabular}
\end{table*}

\begin{table*}[]
\vspace{-2pt}
\footnotesize
\centering
\setlength{\tabcolsep}{5pt} 
\caption{Comparison of fine-tuning strategies on SmolVLM-500M for Text-to-OSM on the CVG-Text Benchmark.}
\label{tab:fine-tuning_osm}
\resizebox{\textwidth}{!}{
\begin{tabular}{@{}l|cccc|cccc|cccc@{}} 
\toprule
\multirow{3}{*}{\textbf{backbone}} & \multicolumn{12}{c}{\textbf{Ground Text $\to$ OSM Image}} \\
\cmidrule(lr){2-13} 
& \multicolumn{4}{c|}{\textbf{New York}} & \multicolumn{4}{c|}{\textbf{Brisbane}} & \multicolumn{4}{c}{\textbf{Tokyo}} \\
\cmidrule(lr){2-5} \cmidrule(lr){6-9} \cmidrule(lr){10-13}
& \textbf{R@1} & \textbf{R@5} & \textbf{L@50} & \textbf{L@100} & \textbf{R@1} & \textbf{R@5} & \textbf{L@50} & \textbf{L@100} & \textbf{R@1} & \textbf{R@5} & \textbf{L@50} & \textbf{L@100} \\
\midrule
SmolVLM-500M (Full) & 41.58 & 64.83 & 45.42 & 49.17 & 38.83 & 62.25 & 43.67 & 46.50 & 22.67 & 46.33 & 28.42 & 35.17 \\
SmolVLM-500M (LoRA) & 53.92 &76.42 & 60.00 & 63.83& 40.33 &64.83 & 45.25 &48.92 & 29.17 &54.92& 36.58&42.50 \\
\textbf{Gain} & \small{$+12.34$} &\small{$+11.59$}&\small{$+14.58$} &\small{$+14.66$} &\small{$+1.5$} &\small{$+2.58$} &\small{$+1.58$}&\small{$+2.42$} &\small{$+6.5$}&\small{$+8.59$}&\small{$+8.16$}&\small{$+7.33$} \\

\bottomrule
\end{tabular}}
\end{table*}

\begin{table}[]
\vspace{-5pt}
\footnotesize
\centering
\setlength{\tabcolsep}{7pt} 
\caption{Analysis of LoRA Scaling Factor ($\alpha$) on on Text-to-Satellite task of CVG-Text New York subset. The LoRA rank ($r$) is fixed at 16, and the scaling factor $\alpha$ is varied. [Key: \textbf{Best}]}
\label{tab:lora_setting}
\begin{tabular}{@{}lc|cccccc@{}} 
\toprule
$\frac{\alpha}{r}$ & $\mathbf{\alpha}$ & \textbf{R@1} & \textbf{R@5}  & \textbf{L@50} & \textbf{L@100} \\
\midrule
1 & 16 & 29.08 & 66.5 & 31.58 & 35.83 \\

2 & 32 & 34.00 & 69.92  & 36.5 & 40.58  \\

4 & 64 & 39.25 & 75.75 & 41.75 & 45.17 \\

\textbf{8} & 128 & \textbf{44.75} &  \textbf{75.92}  & \textbf{47.50} & \textbf{51.33} \\
\bottomrule
\end{tabular}
\vspace{-15pt}
\end{table}

\begin{table}[]
\footnotesize
\centering
\setlength{\tabcolsep}{7pt} 
\caption{Analysis of the loss temperature($\tau$) on Text-to-Satellite task of CVG-Text New York subset. [Key: \textbf{Best}]}
\label{tab:temperature_ablation}
\begin{tabular}{@{}l|cccccc@{}}
\toprule
\textbf{$\tau$} & \textbf{R@1} & \textbf{R@5} & \textbf{L@50} & \textbf{L@100}  \\
\midrule
0.02 & 41.33 & 74.33  & 44.66 & 48.00  \\
\textbf{0.03} & \textbf{44.75} & \textbf{75.92} & \textbf{47.50} & \textbf{51.33} \\
0.05 & 41.83 & 75.75 & 44.58 & 48.50  \\
0.07 & 39.25 & 75.75 & 41.75 & 45.17  \\
0.1 &  37.42 & 72.08 & 39.67 & 43.25  \\
Learnable & 39.08& 75.92 & 41.58 & 44.92 \\

\bottomrule
\end{tabular}
\vspace{-15pt}
\end{table}

\subsection{Fine-tuning Study}
\label{subsec:fine-tuning}
To study our \textbf{RQ3} regarding the optimal fine-tuning strategy for NGCG, we first compare Full Fine-Tuning (FFT) with LoRA fine-tuning for the Text-to-Satellite task across three cities by using SmolVLM-500M and InternVL3.5-1B. The results are shown in \Cref{tab:fine-tuning_satellite}. As InternVL3.5-1B, pre-trained on massive data, already possesses highly robust features, it achieves a higher base performance. Compared with that, SmolVLM-500M exhibits a notable improvement under LoRA across all three cities. This is evidence that narrower representation is highly adaptable and benefits significantly from LoRA's regularization, which prevents overfitting. We also conduct fine-tuning strategy experiments for the Text-to-OSM task in \Cref{tab:fine-tuning_osm}. The smaller percentage gain on the Text-to-OSM task, when compared to the Text-to-Satellite task, suggests that OSM images are easier for the MLLM. Because OSM images inherently contain more explicit details (e.g., bus stops, store names), the MLLM’s existing pretrained knowledge is readily applicable, resulting in less room for dramatic improvement via LoRA fine-tuning. We observe that data quality affects LoRA fine-tuning performance. For example, due to the limited data quality in Brisbane OSM data \cite{cvgtext}, LoRA fine-tuning shows a relatively small improvement on it. LoRA fine-tuning demonstrates enhanced generalization and robustness to data scarcity (\textit{see Supplementary for details}).

\noindent\textbf{Takeaway:} LoRA serves as the optimal fine-tuning strategy for NGCG. It not only provides parameter efficiency but also acts as a regularizer for the NGCG task, especially when utilizing a smaller-scale MLLM.

\begin{figure*}[]
\vspace{-15pt}
    \centering
    \includegraphics[width=.78\linewidth]{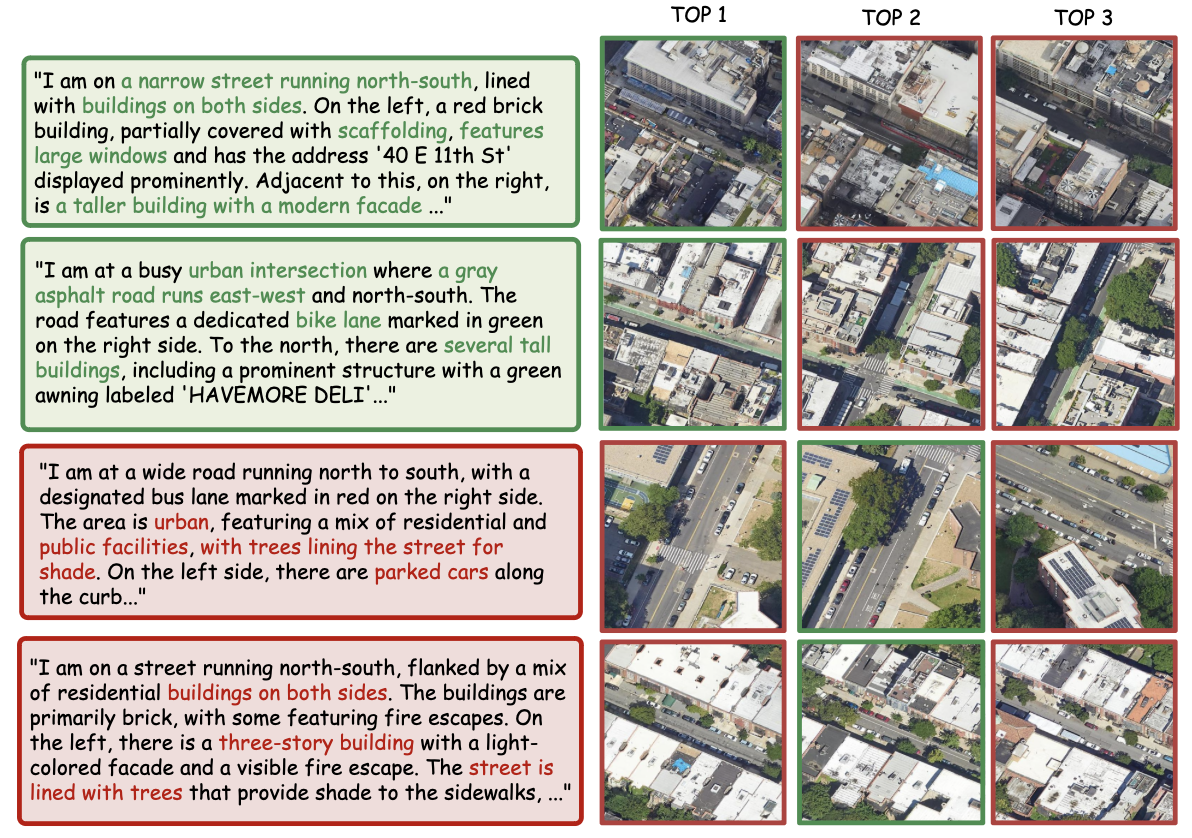} 
    \vspace{-10pt}
    \caption{Visualization of Text-to-Satellite Image Retrieval Results on CVG-Text. Four text-satellite pairs are shown. Left is the query text. On the right is the the retrieved satellite images. Ground Truth satellite images are outlined with a green border, while incorrect matches are outlined with a red border.}
    \label{fig:vis}
    \vspace{-10pt}
\end{figure*}

\subsection{Effect of LoRA Setting}
\label{subsec:lora}
It is also curious that how the LoRA configuration influences the geo-localization accuracy, that is centralized in our \textbf{RQ4}. To study this, we fix LoRA rank $r$ to 16 and vary the $\frac{\alpha}{r}$ ratio to four values, 1, 2, 4, and 8. As shown in the \Cref{tab:lora_setting}, increasing the scaling factor $\alpha$ leads to significant improvements across all retrieval metrics (R@1, R@5, L@50, and L@100). The highest score $44.75\%$ at R@1 is when $\alpha$ = 128, corresponding to the highest ratio ($\frac{\alpha}{r}=8$). This trend suggests that for NGCG task, a higher scaling factor, which controls the magnitude of the weight update $\Delta W$, is necessary to effectively adapt the pre-trained MLLM to the new task. A smaller $\alpha$ results in overly constrained updates, which limits the model's ability to learn the necessary task-specific cross-modal alignments. 

\noindent\textbf{Takeaway:} 
A higher value of $\alpha$ is necessary for the NGCG task because it directly controls the magnitude of the weight update $\Delta W$ during fine-tuning, which is required to effectively adapt the pre-trained MLLM's features to the specific cross-modal alignments needed for NGCG.





\subsection{Learning Objective Sensitivity}
Finally, we conduct a sensitivity analysis on the temperature hyperparameter ($\tau$), which directly controls the sharpness of the contrastive loss to study our \textbf{RQ5}. As shown in \Cref{tab:temperature_ablation}, setting the temperature to a fixed value of $\tau=0.03$ yields the best performance. A relatively low temperature value ($\tau=0.02$) leads to an overly strict separation between features, potentially causing training instability or hindering convergence due to high variance in gradient estimation. On the other hand, a high value ($\tau=0.1$) uniformly increases the probability contributed by all negative samples. This effectively degrades the gradient of hard negatives within each mini-batch. Consequently, the model receives less informative gradients, which prevents it from learning the fine-grained distinctions necessary for accurate cross-modal alignment in NGCG. Interestingly, using a learnable temperature, which is widely adopted in prior works~\cite{sample4geo,cvgtext,ye2024cross}, also underperforms compared to the fixed setting of $\tau=0.03$. This suggests that allowing the temperature to evolve during training may not stabilize the contrastive loss, or it might converge to a sub-optimal value.

\noindent\textbf{Takeaway:} A finely-tuned fixed temperature ($\tau=0.03$) is necessary to properly regulate the feature distribution, preventing the model converge into a sub-optimal.

\subsection{Inference Efficiency and Practical Latency}

As shown in \Cref{tab:computational_analysis}, our models establish a new performance-efficiency frontier for NGCG. While the use of MLLM backbones introduces higher latency, this is a necessary trade-off to achieve the accuracy gains (+10.7\% R@1). We emphasize that GPU memory is the hard constraint for deployment. Ours-S requires only $\sim$683 MB, allowing it to run on edge devices where high-memory models cannot even initialize. While we currently prioritize reasoning accuracy, the observed latency can be effectively mitigated in future through compression techniques like quantization and knowledge distillation.

\begin{table}[!h]
\setlength{\tabcolsep}{2pt}
\footnotesize
\centering
\caption{Efficiency Analysis. The computational cost of Ours-S with SmolVLM-256M and Ours-I with InternVL3.5-1B against two dual-encoder baselines. Bold values indicate the most efficient performance in each category. }
\vspace{-5pt}
\begin{tabular}{l|cccc}

\toprule
\textbf{Method}   &\textbf{GFLOPs (G)} & \textbf{Latency (ms)}  & \textbf{Memory (MB)} \\
\midrule
GeoText &  \textbf{49.34 }& \textbf{9.64} & 1268.54\\ 
CrossText2Loc&200.25  & 20.22& 2979.86 \\ 
\midrule
\textbf{Ours-S} & 130.24 &33.28 &\textbf{682.98}\\
\textbf{Ours-I} & 581.35 &100.44 &2258.85 \\
\bottomrule
\end{tabular}
\label{tab:computational_analysis}
\end{table}

\vspace{-8pt}
\subsection{Qualitative Visualization}
We randomly select four retrieval examples from CVG-Text in \Cref{fig:vis}. The green border indicates ground truth satellite images. In the first example, the model achieves precise localization by learning to combine architectural details (e.g., 'scaffolding', 'large windows') with spatial cues (e.g., 'narrow street running north–south'). Similarly, the second example shows robustness in ambiguous urban environments by emphasizing distinctive features like the 'gray asphalt road' and 'bike lane', disambiguating visually similar intersections. The last two examples highlight failure cases where the model does not achieve a Top-1 rank, as indicated by the red outlines. Although our method correctly incorporates the spatial information from the queries, the similar visual characteristics shared by all candidate images make precise ranking difficult. For example, the third query depends on general contextual descriptions (e.g., 'urban,' 'public facilities,' 'trees lining the street'). Likewise, in the final example, the description consists of elements that are common to all candidates (e.g., 'buildings on both sides,' 'street is lined with trees'). The results demonstrate that our method enables strong alignment and robust understanding across varied descriptive styles and visual complexities. \textit{ More examples refer to the supplementary.}

%% file: sec/5_conclusion.tex
\subsection{Conclusion}
\label{subsec:conclusion}
In this work, we are the first to introduce Multimodal Large Language Models (MLLMs) as feature encoders for the challenging Natural-language Guided Cross-view Geo-localization (NGCG) task. Our framework effectively overcomes the design complexity of prior approaches while excelling at cross-modal feature alignment. It achieves SOTA performance on GeoText-1652 and competitive results on CVG-Text. Through a comprehensive study guided by five research questions, we provide systematic insights and practical guidelines for adapting MLLMs to NGCG. Our findings reveal that parameter-efficient adaptation effectively adapts rich pretrained knowledge for cross-modal localization, laying the groundwork for future exploration and innovation in the field of NGCG.